\pdfoutput=1
\documentclass[11pt,a4paper]{article}
\usepackage[hyperref]{acl2021}
\usepackage{times}
\usepackage{latexsym}
\usepackage{graphicx}
\usepackage{amsmath}

\DeclareUnicodeCharacter{043D}{ }
\DeclareUnicodeCharacter{0410}{ }
\DeclareUnicodeCharacter{0442}{ }
\DeclareUnicodeCharacter{0440}{ }
\DeclareUnicodeCharacter{043E}{ }
\DeclareUnicodeCharacter{043F}{ }
\DeclareUnicodeCharacter{043B}{ }
\DeclareUnicodeCharacter{0433}{ }
\DeclareUnicodeCharacter{0456}{ }
\DeclareUnicodeCharacter{0447}{ }
\DeclareUnicodeCharacter{0432}{ }
\DeclareUnicodeCharacter{0438}{ }
\DeclareUnicodeCharacter{043C}{ }
\DeclareUnicodeCharacter{0444}{ }
\DeclareUnicodeCharacter{0441}{ }
\DeclareUnicodeCharacter{044C}{ }
\DeclareUnicodeCharacter{043A}{ }
\DeclareUnicodeCharacter{0445}{ }
\DeclareUnicodeCharacter{0434}{ }
\DeclareUnicodeCharacter{0436}{ }
\DeclareUnicodeCharacter{0435}{ }
\usepackage[utf8]{inputenc}

\usepackage{tabularx,booktabs}
\newcolumntype{C}{>{\centering\arraybackslash}X} % centered version of "X" type
\usepackage{microtype}

\aclfinalcopy % Uncomment this line for the final submission
\title{Transfer Learning based Speech Affect Recognition in Urdu}

\author{Sara Durrani \\
  Fast Nuces \\
  Islamabad, Pakistan.\\
  \texttt{sara.durrani@nu.edu.pk} \\\And
  Muhammad Umair Arshad \\
  Fast Nuces \\
  Islamabad, Pakistan.\\
  \texttt{umair.arshad@nu.edu.pk} \\}

\date{}

\begin{document}
\maketitle
\begin{abstract}
It has been established that Speech Affect Recognition for low resource languages is a difficult task. Here we present a Transfer learning based Speech Affect Recognition approach in which: we pre-train a model for high resource language affect recognition task and fine tune the parameters for low resource language using Deep Residual Network. Here we use standard four data sets to demonstrate that transfer learning can solve the problem of data scarcity for Affect Recognition task. We demonstrate that our approach is efficient by achieving 74.7 percent UAR on RAVDESS as source and Urdu data set as a target. Through an ablation study, we have identified that pre-trained model adds most of the features information, improvement in results and solves less data issues. Using this knowledge, we have also experimented on SAVEE and EMO-DB data set by setting Urdu as target language where only 400 utterances of data is available. This approach achieves high Unweighted Average Recall (UAR) when compared with existing algorithms.
\end{abstract}

\section{Introduction}

Speech Affect Analysis and Recognition is an open research problem that is making
an impact in the research community due to its applications. Different psychologists are doing researches to identify the emotions of human being and their interdependence \citet{bornstein2006complex}. This work leads to the identification of correct affects and their influence on society due
to human behaviours. The human actions are the result of their thoughts generated due to affects (emotions) \citet{pavlova2018phenomenon}. Human Computer Interaction (HCI) \citet{carroll2001evolution} contains multiple applications of Speech Affect Recognition (SAR). The speech affect contains applications like
customer shopping experience \citet{kawaf2017construction}, smart cities \citet{roza2016citizen}, smart classrooms \citet{shao2019emotions} , smart environment \citet{huang2019speech} and many more. These affects include sadness, anger, disgust, happiness, and many more. In the exploration of application’s dependency on speech affects we have found that human behaviour can be tracked using the Automatic Speech Affect Recognition System. It is used to monitor human affects in call centers with the customers \citet{vidrascu2005detection}. These systems are also being used to track depression and mental health in different
smart homes and offices \citet{basu2017review}. \\

The Automatic Speech Affect Recognition systems work
by identifying different speakers and their emotions. The models developed can be
speaker dependent or independent. The problem arises when different models have to
deal with speech data that comes from a variety of speakers from different age groups,
gender, accent, and language. In this context, language is a huge barrier for a lot of
speech recognition systems. Speech includes different affects that make a huge impact
on the performance of Speech Affect Recognition Systems. In recent years, many methods have been proposed to solve the problems related to
Speech Affect Recognition. In paper \cite{seo2020fusing} has extracted the feature vector using bag of visual words. The bag of visual words has assisted to learn local and global features in log-Mel spectrogram by constructing frequency histogram. The Siamese neural network's loss is modified in \citet{DBLP:conf/aaai/FengC20} for training in transfer learning settings. The results are achieved by the distance loss between the same and different classes. the \cite{liu2020cross} has handled the problem of cross-corpus speech affect recognition as the architecture consists of two modules: first model features are extracted and domain adaptive layer is introduced. Further, the author has introduced the cross-corpus evaluations. The specific work for cross-corpus settings including Urdu has been done by \citet{latif2018cross}. Further, \citet{munot2019emotion} have proved that emotions impact speech recognition. In \citet{latif2018transfer}, the author has worked to improve the classification accuracy using transfer learning in Deep belief Networks.\\

In this work, we are proposing Transfer Learning based Deep Neural Networks that
would play a significant role in Speech Affect features extraction. The baseline Deep
Neural network model plays role as feature extraction tool for identification of Speech
Affects in high resource languages. Further, the transfer learning is applied on Deep Residual Network
to handle the issue of data scarcity. We have used the concept of transfer learning for Urdu language to advance Speech Affect Classification. The contributions of this work in Speech Affect Recognition are as follows:-
\begin{itemize}
    \item The baseline Deep neural network is used as feature extraction tool for more relevant results.
    \item The achievement of transfer learning in Deep Residual Network from English to Urdu and German to Urdu, respectively.

\end{itemize}

\section{Method}

For the Speech Affect Recognition, we use a Deep Neural Network Resnet34 adapted from \citet{he2016deep} as shown in figure 2. We use this architecture for all of the models including baseline and transferable parameters model. We have made sure to fix it for one GPU implementation which helped in reducing the computing resources. \\
We have trained a baseline Resnet34 model for the recognition of Speech Affects on the English data set. The model gets the mel-spectrograms as the input of the audio clips from the feature extraction library Librosa. During the training of the model, the parameters are updated. During the re-training of the whole model by adding a new layer of low resource data, the new and old model shares the same speech features.\\
For the transfer learning settings, we use the pre-train English model by freezing the previous layer and adding the next layer for training in target low resource language. In the end, the final layer has the activation function softmax that generate affect labels for the required speech. This method of transfer learning seems easy and flexible. The one advantage of this method is that the pre-processing steps remain the same for all the language data.

\section{Experimental Setup}
\subsection{Data Sets}

\textbf{RAVDESS.}
We use the RAVDESS multimodal data set \citet{livingstone2018ryerson} of emotional speech and song. The accent of 24 actors is neutral North American English. The database contains gender-balanced statements of emotions that include calm, sad, happy, angry, and fearful. We have only used the speech recordings for our training. The complete set contains 7356 recordings of different emotions. This data set is tested ten times to ensure intensity, correctness, and relevancy.\\
\\
\textbf{URDU Data set.}
We have used Urdu data set \citet{latif2018cross} for cross corpus evaluations and training. This data set contains four baseline emotions neutral, anger, happy and sad. The utterances are from 3 speakers including 27 male and 11 female. This data set contains 400 speech utterances.\\
\\
\textbf{SAVEE.}
We use SAVEE \citet{jackson2014surrey} for baseline training of our English Speech Affects model. This database is recorded by four male graduates who are English natives. It consists of 120 utterances per person. This data set targets speech affects that include fear, anger, happiness, disgust, sadness, and surprised. The speech data labelling has been done by automatic labelling, Speech Filling System and manual resources.\\
\\
\textbf{Emo-DB.}
Emo-DB \citet{burkhardt2005database} is a German language data set that is recorded by 10 actors. These actors include five males and five females. The recorded emotions are neutral, fear, anger, joy, sadness, disgust, and boredom. These utterances are further tested by 20 persons to verify each emotion. We have used this data set to train baseline German model to evaluate cross corpus on the Urdu language.

\subsection{Preprocessing}
\textbf{Speech.} The preprocessing of data includes the arrangement of shuffled data under correct labels of affects. Each speaker's recordings are arranged for all speech labels in the respective folders. Further, the voice clips are used to extract the mel-spectogram and MFCCs features. These features are extracted on an individual speaker level are independent of any hindrance or noise.

\subsection{Model Architecture for SAR}
\textbf{Residual Deep CNN Network.}
The deep CNN networks are solving many general problems. As shown in figure 2, the model is composed of residual blocks that are stacked onto each other. Each of the block  is connected directed with the upper and lower layer. The following equation defines a single block of ResNet according to Figure 1 given:-\\

\begin{align*}
\centering
  b= f(a,w_i)+a
\end{align*}

where a is the input and b is the output of the layer. We consider F as the non-linear function. The training speed has improved a lot due to the existence of residual connections between blocks. The dropout rate is set to 0.1. The architecture also gets better results with an increase in network depth.\\
\\

\begin{figure}[h]
    \centering
    \includegraphics[width=0.2\textwidth]{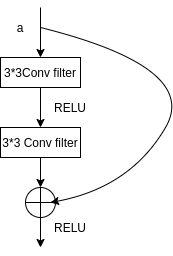}
    \caption{This figure is showing the structure of the block of Resnet34 \citet{he2016deep}. The filters are of size 3*3 in each layer and skip connections is mapping a.}
    \label{fig:mesh1}
\end{figure}

\begin{figure}[h]
    \centering
    \includegraphics[width=0.2\textwidth]{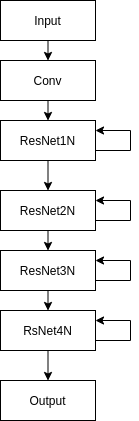}
    \caption{Network architecture of ResNet with arrow lines that represent residual connections.}
    \label{fig:mesh1}
\end{figure}

\begin{table*}
 \caption{Details about network layers ResNet34}
\label{my-label}
\begin{tabularx}{\textwidth}{@{}l*{4}{C}c@{}}
\toprule
Layers     & Output size & [filter,\#map*width] & stride \\ 
\midrule
conv1   & 112*112      & [7*7,64]         & 2   \\ 
ResnetN1     & 56*56        & [3*3,64]*3, [3*3,64]*3        & (2,2)     \\ 
ResnetN2      & 28*28       & [3*3,128]*4,[3*3,128]*4          & (2,2)  \\ 
ResnetN3      & 14*14     & [3*3,256]*6,[3*3,256]*6          & (2,2)   \\ 
ResnetN4       & 7*7       & [3*3,512]*3,[3*3,512]*3           & (2,2)  \\ 
&               &averagepool, Softmax Layer\\
Flops&       &$3.6*10^9 $        \\ 
\bottomrule
\end{tabularx}
\end{table*}

\textbf{Implementation and Training.}We have adopted the model from \citet{he2016deep} ResNet34 by using four blocks. These blocks are named BlockN1, BlockN2, BlockN3, and BlockN4 as shown in the figure. The filters used are of size 7*7,64. Table 1 shows the architecture for layers of details of ResNet. Each network layer of CNN is stride by a factor of 2. We have also applied RELU activation \citet{nair2010rectified} and softmax. The batch normalization \citet{ioffe2015batch} is also used in the network after every convolution before the activation for reduction of the shift in covariance. The learning rate of 0.0001 is used in the PyTorch framework. First, the model is trained from scratch on the English data set. Then, it is freeze and then the last layer is trained from scratch in the pytorch framework. In the end, the whole model has trained again for a total of 100 epochs. The batch size is set up to 25 and the training testing ratio is kept 80 and 20, respectively. 

\subsection{Evaluation Metrics}
\textbf{Metrics.} We discuss the UAR(unweighted average recall) for all of our experiments. High UAR is difficult to achieve as it is directly affected by the performance of the model. In low resource settings, mostly UAR scores tend to be low. Unweighted Average Recall is a parameter that is calculated for the recall of every class. It gives out an easy calculation for accuracy when the data set samples count is imbalanced as compared to all other classes.
\section{Results}
\subsection{Baseline SAR Results}
By using the information and experimental setup of section 3, we have trained baseline models in English and German. We have reported the UAR scores in the table for each model. Our goal is to make use of the pre-trained model for the transfer learning settings. We have stopped the training of the baseline model after 2 epochs then we have done training in transfer learning settings. We have found that better pre-training of the baseline model will yield better transfer learning results.\\
\\
\begin{tabular}{l*{6}{c}r}
              & RAVDESS & SAVEE & Emo-DB \\
\hline
UAR           & 62.65 & 70.61    & 84.79 \\
\end{tabular}

\subsection{Multilingual Settings}
\textbf{English to Urdu SAR Evaluations.} We have used the name of the model as Eng-Urdu to identify the source to target language training. The graph and table show the UAR achieved for English to Urdu model performance. The baseline model results improve substantially as compared to the existing results. The reason is that transfer learning has contributed a lot to the transfer of features from high resource language to low resource language. The transfer learning has improved results from 62.65 to 74.7 UAR for RAVDESS and from 72.61 to 84.18 for SAVEE.\\

Overall the baseline model struggled due to fewer data to achieve high recall and precision. With transfer learning, we have observed gains in UAR of about 10 to 15 points. The fine-tuning of the SAR model can somehow produce similar results but it requires three times more of the data which is a difficult task. The ideal UAR results with small data sets are easy to achieve in limited time scenarios. The performance benchmark data and time limitations have been satisfied by our proposed approach. \\
\\
\textbf{German to Urdu Evaluations.} For this setting, we have trained the baseline German model on the Emo-DB data set. We have trained the baseline for 15 epochs with a learning rate of 0.00001. A new layer of Urdu data has been added that has improved the overall UAR of the model. The transfer learning settings have solved the problem of data scarcity for Urdu. Therefore, we have achieved the UAR of 87.5 percent for German-Urdu Speech Affect Recognition.

\begin{tabular}{l*{3}{c}r}
              & UAR  \\
\hline
\citet{latif2018cross}              \\
EMO-DB                               & 57.87\\
SAVEE                               & 40.10\\

\hline
RESNet34 SAR(proposed)               \\
RAVDESS                         &74.7 \\
SAVEE                           & 84.18\\
Emo-DB                          &87.5\\

\end{tabular}

\section{Analysis}
From the experiments, transfer learning stands out in terms of achieving high UAR scores given in Figure 3. This makes us understand that training models under multilingual settings in scenarios of scarce data gain good scores and learn many different features from multiple languages. We have learned that our transfer learning-based Speech Affect Recognition solves the problems of time and data limitations. In addition to this, it has also added a transfer learning-based approach for low resource languages. \\

Another aspect that we have learned is that pre-trained models add towards the improved accuracy and UAR. It learns transferable low-level acoustic features that reduce the problem of speakers and channel differences. In the end, the pre-trained model adds information in the training of low resource language. Therefore, these experiments have shown the advantage of multilingual, cross-corpus, and transfer learning training settings specific to low resource language like Urdu.

\begin{figure}[h]
    \centering
    \includegraphics[width=0.5\textwidth]{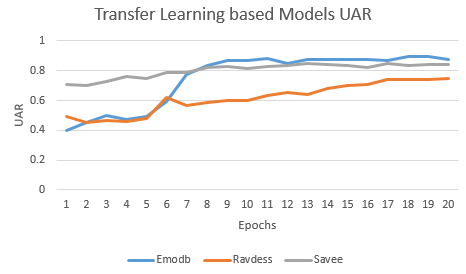}
    \caption{UAR for Transfer learning based models trained on Emo-DB, Ravdess and SAVEE tested on Urdu data set.}
    \label{fig:mesh1}
\end{figure}

\section{Conclusion}
This paper discusses the idea of pre-training a Deep Residual Network ResNet34 for low resource language that is Urdu. We showed that high UAR can be achieved by training the baseline model for high resource languages (English, German) and applying transfer learning for low resource language. We achieved 74.7 UAR for English-Urdu SAR and 87.5 UAR for German-Urdu SAR. We identified that the pre-trained model captured more acoustic features and contributed more towards low resource language. From these findings, practical applications can be developed for Speech Affect Recognition using data from multiple languages. This approach made speech affect recognition easy with less or no data set.

\bibliography{acl2021}
\bibliographystyle{acl_natbib}

\end{document}